\documentclass[11pt,a4paper]{article}
\usepackage{hyperref}
\usepackage[hyperref]{acl2020}
\usepackage{times}
\usepackage{latexsym}

\usepackage{microtype}

\usepackage{url}
\usepackage{times}
\usepackage{latexsym}
\usepackage{multirow}
\usepackage{amssymb}
\usepackage{pifont}
\newcommand{\cmark}{\ding{51}}%
\newcommand{\xmark}{\ding{55}}%
\usepackage{url}
\usepackage{enumitem}
\usepackage{multirow}
\usepackage{graphicx}
\usepackage{comment}
\usepackage{hhline}
\usepackage{longtable}

\usepackage{titlesec}
\titlespacing*{\section}{0pt}{0.45\baselineskip}{0.45\baselineskip}

\newcommand{\pgn}{PGN}
\newcommand{\pgnp}{PGN-Pre}
\newcommand{\dgpt}{D-GPT}
\newcommand{\gpt}{GPT-2}
\newcommand{\gptp}{GPT-2-Pre}

\newcommand{\coqa}{CoQA}
\newcommand{\quac}{QuAC}
\newcommand{\squadtwo}{SQuAD 2.0}
\newcommand{\glove}{GloVe}
\newcommand{\elmo}{ELMo}
\newcommand{\logistic}{\textbf{Logistic}}
\newcommand{\softmax}{\textbf{Softmax}}
\newcommand{\quotes}[1]{``#1''}
\newcommand{\lgr}{Syntactic Transformations}
\newcommand{\brc}{best response classifier}
\newcommand{\SeqToSeq}{\textsc{Seq2Seq}}

\aclfinalcopy 
\def\aclpaperid{1469} 

\title{Fluent Response Generation for Conversational Question Answering}

\author{Ashutosh Baheti, Alan Ritter \\
  Computer Science and Engineering \\
  Ohio State University \\
  \texttt{\{baheti.3, ritter.1492\}\@osu.edu} \\\And
  Kevin Small \\
  Amazon Alexa \\
  \texttt{smakevin@amazon.com} \\}

\date{}

\begin{document}
\maketitle

\begin{abstract}

Question answering (QA) is an important aspect of open-domain conversational agents, garnering specific research focus in the conversational QA (ConvQA) subtask. One notable limitation of recent ConvQA efforts is the response being answer span extraction from the target corpus, thus ignoring the natural language generation (NLG) aspect of high-quality conversational agents. In this work, we propose a method for situating QA responses within a {\SeqToSeq} NLG approach to generate fluent grammatical answer responses while maintaining correctness. From a technical perspective, we use data augmentation to generate training data for an end-to-end system. Specifically, we develop {\lgr} (STs) to produce question-specific candidate answer responses and rank them using a BERT-based classifier \citep{devlin-etal-2019-bert}. Human evaluation on {\squadtwo} data \citep{rajpurkar-etal-2018-know} demonstrate that the proposed model outperforms baseline {\coqa} and {\quac} models in generating {\em conversational} responses. We further show our model's scalability by conducting tests on the {\coqa} dataset.\footnote{The code and data are available at \href{https://github.com/abaheti95/QADialogSystem}{https://github.com/abaheti95/QADialogSystem}.}
\end{abstract}

\section{Introduction}
\label{sec:intro}
Factoid question answering (QA) has recently enjoyed rapid progress due to the increased availability of large crowdsourced datasets (e.g., SQuAD \citep{rajpurkar-etal-2016-squad}, MS MARCO \citep{bajaj2016ms}, Natural Questions \citep{kwiatkowski-etal-2019-natural}) for training neural models and the significant advances in pre-training contextualized representations using massive text corpora (e.g., ELMo \citep{peters-etal-2018-deep}, BERT \citep{devlin-etal-2019-bert}).  Building on these successes, recent work examines {\em conversational} QA (ConvQA) systems capable of interacting with users over multiple turns.  Large crowdsourced ConvQA datasets (e.g., {\coqa} \citep{reddy-etal-2019-coqa}, {\quac} \citep{choi-etal-2018-quac}) consist of dialogues between crowd workers who are prompted to ask and answer a sequence of questions regarding a source document.  
Although these ConvQA datasets support multi-turn QA interactions, the responses have mostly been limited to extracting text spans from the source document and do not readily support abstractive answers \citep{yatskar-2019-qualitative}.
While responses copied directly from a Wikipedia article can provide a correct answer to a user question, they do not sound natural in a conversational setting.  To address this challenge, we develop {\SeqToSeq} models that generate fluent and informative answer responses to conversational questions.

To obtain data needed to train these models, rather than constructing yet-another crowdsourced QA dataset, we transform the answers from an existing QA dataset into fluent responses via data augmentation.
Specifically, we synthetically generate supervised training data by converting questions and associated extractive answers from a SQuAD-like QA dataset into fluent responses via \textit{\lgr} (STs).
\begin{figure*}[h]
    \centering
    \includegraphics[width=\textwidth]{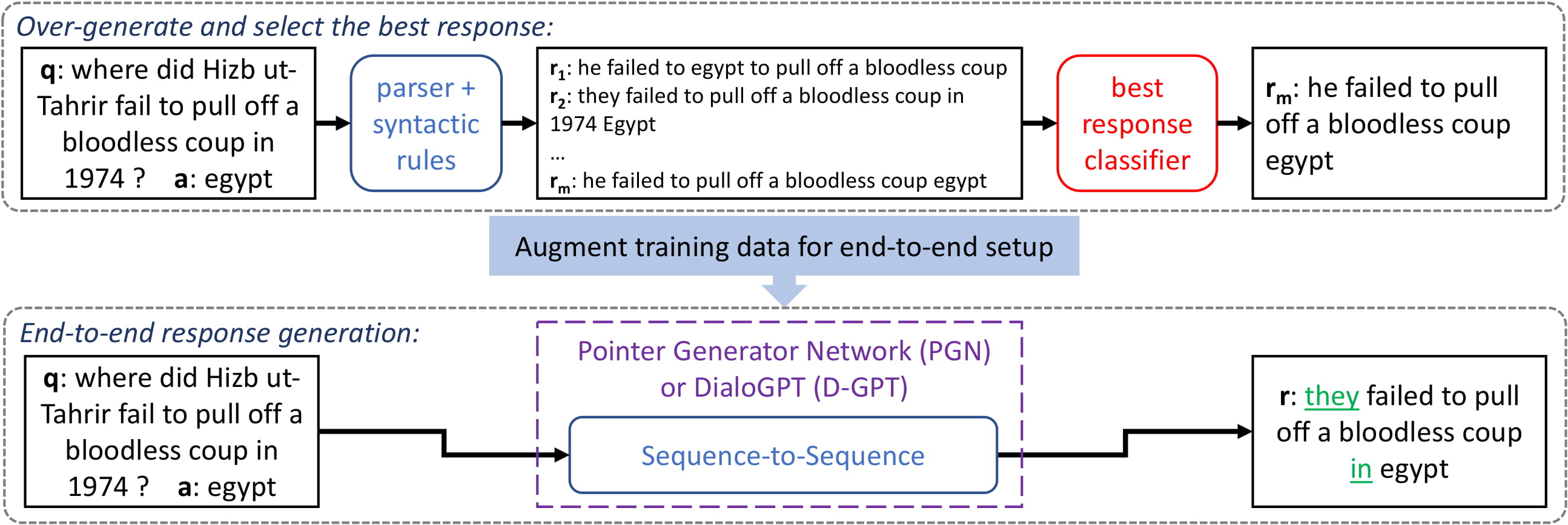}
    \caption{Overview of our method of generating conversational responses for a given QA. In the first method, the \textit{\lgr} (STs) over-generate a list of responses (good and bad) using the question's parse tree and the \textit{\brc} selects the most suitable response from the list. Our second method uses this pipeline to augment training data for training a {\SeqToSeq} networks PGN or D-GPT (\S \ref{subsec:models}). The final {\SeqToSeq} model is end-to-end, scalable, easier to train, and performs better than the first method exclusively.}
    \label{fig:overview}
    \vspace{-4mm}
\end{figure*}
These STs over-generate a large set of candidate responses from which a BERT-based classifier selects the best response as shown in the top half of Figure~\ref{fig:overview}. 

While over-generation and selection generates fluent responses in many cases, the brittleness of the off-the-shelf parsers and the syntatic transformation rules prevent direct use in cases that are not well-covered.  To mitigate this limitation, we generate a new augmented training dataset using the best response classifier that is used to train end-to-end response generation models based on Pointer-Generator Networks (PGN) \citep{see-etal-2017-get} and pre-trained Transformers using large amounts of dialogue data, DialoGPT (D-GPT) \cite{zhang2019dialogpt}.  
In \S \ref{sec:squad_experiments} and \S \ref{sec:cross_domain_experiments}, we empirically demonstrate that our proposed NLG models are capable of generating fluent, abstractive answers  on both {\squadtwo} and \coqa.

\section{Generating Fluent QA Responses}
In this section, we describe our approach for constructing a corpus of questions and answers that supports fluent answer generation (top half of Figure \ref{fig:overview}). We use the framework of \textbf{overgenerate and rank} previously used in the context of question generation~\citep{heilman-smith-2010-good}. We first \textbf{overgenerate} answer responses for QA pairs using STs in \S \ref{sec:rules}. We then \textbf{rank} these responses from best to worst using the response classification models described in \S \ref{sec:response_classifiers}. Later in \S \ref{sec:aug_and_generate}, we describe how we augment existing QA datasets with fluent answer responses using STs and a best response classifier. This augmented QA dataset is used for training the PGN and Transformer models.


\subsection{Syntactic Transformations (STs)}
\label{sec:rules}
\begin{figure}[t]
    \centering
    \includegraphics[width=\linewidth]{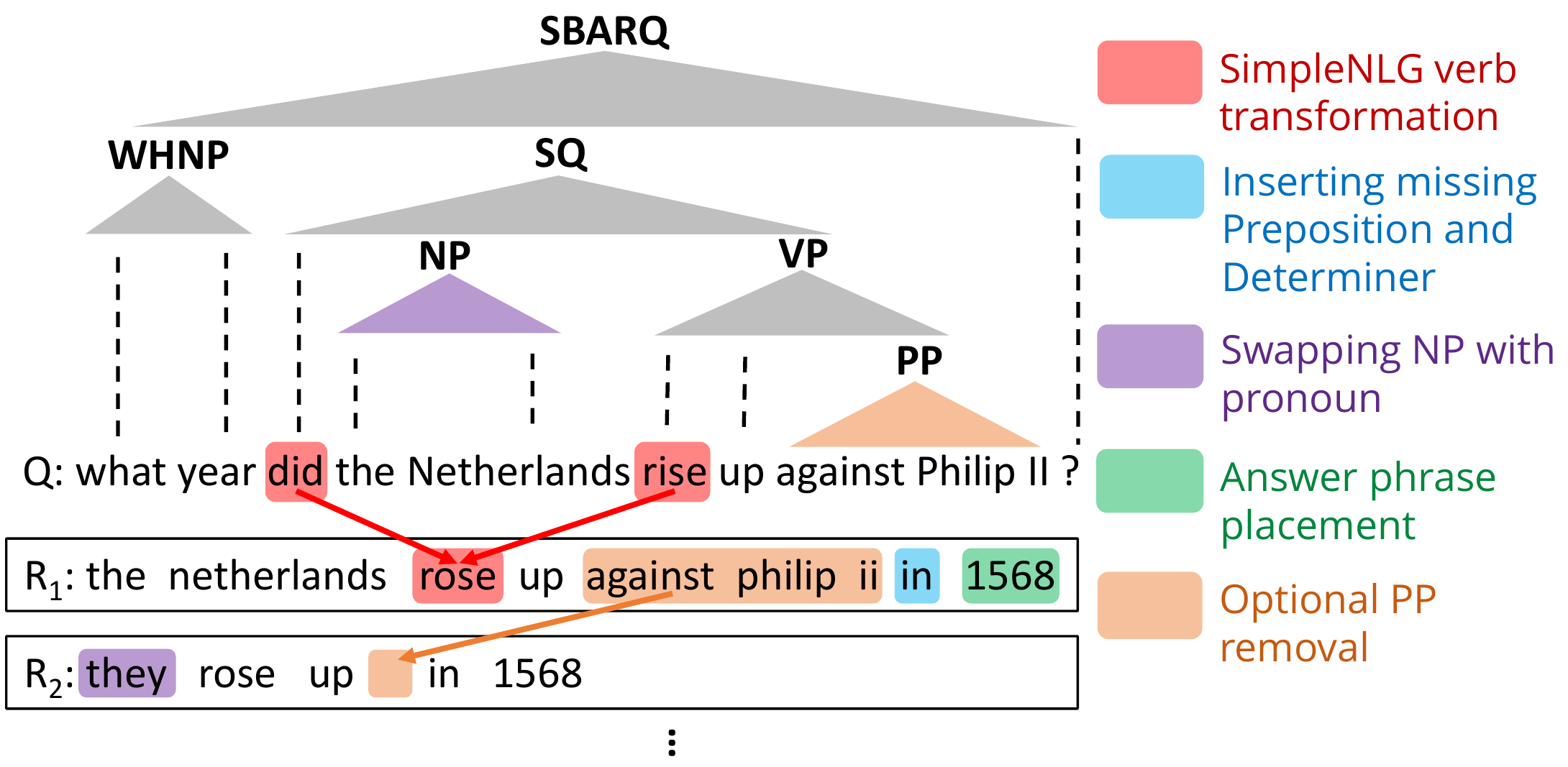}
    \caption{An example of Syntactic Transformations in action. Question: \quotes{what year did the Netherlands rise up against Philip II?} Answer: \quotes{1568}. Using the question's parse tree we: (1) modify the verb \quotes{rise} based on the auxiliary verb \quotes{did} (red); (2) add missing prepositions and determiners (sky blue); (3) combine the subject and other components with the answer phrase (green) to generate the candidate $R_1$. In another candidate $R_2$, we swap the subject with pronoun \quotes{they} (purple). Our transformations can also optionally remove Prepositional-Phrases (PP) as shown in $R_2$ (orange). In the figure, we only show two candidates but in reality the transformations generate many more different candidates, including many implausible ones.}
    \label{fig:linguistic_transformation}
\end{figure}

The first step is to apply the Syntactic Transformations (STs) to the question's parse tree along with the expert answer phrase to produce multiple candidate responses. For the STs to work effectively accurate question parses are essential. We use the Stanford English lexparser\footnote{\href{https://nlp.stanford.edu/software/parser-faq.html\#z}{https://nlp.stanford.edu/software/parser-faq.html\#z}}\citep{klein2003fast}, which is trained on WSJ sections 1-21, QuestionBank \citep{judge2006questionbank}, amongst other corpora. However, this parser still fails to recognize $\sim 20\%$ of the questions (neither SBARQ nor SQ tag is assigned). For such erroneous parse trees, we simply output the expert answer phrase as a single response. The remaining questions are processed via the following transformations to over-generate a list of candidate answers:
(1) \textbf{Verb modification}: change the tense of the main verb based on the auxiliary verb using SimpleNLG \citep{gatt2009simplenlg}; (2) \textbf{Pronoun replacement}: substitute the noun phrase with pronouns from a fixed list; (3) \textbf{Fixing Preposition and Determiner}: find the preposition and determiner in the question's parse tree that connects to the answer phrase and add all possible prepositions and determiners if missing. (4) \textbf{Response Generation}: Using Tregex and Tsurgeon \citep{levy-andrew-2006-tregex}, compile responses by combining components of all previous steps and the answer phrase. 
In cases where there are multiple options in steps (2) and (3), the number of options can explode and we use the best response classifier (described below) to winnow. An example ST process is shown in Figure \ref{fig:linguistic_transformation}.

\subsection{Response Classification and Baselines}
\label{sec:response_classifiers}
A classification model selects the best response from the list of ST-generated candidates. Given the training dataset, $D$, described in \S \ref{sec:dataset} of $n$ question-answer tuples $(q_i, a_i)$, and their list of corresponding responses, $\{r_{i1}, r_{i2}, ... , r_{im_i}\}$, the goal is to classify each response $r_{ij}$ as bad or good. The probability of the response being good is later used for ranking. We experiment with two different model objectives described below,

\noindent
\logistic: We assume that the responses for each $q_i$ are independent of each other. The model ($F()$) classifies each response separately and assigns 1 (or 0) if $r_{ij}$ is a good (or bad) response for $q_i$. The {\logistic} loss is given by $\sum_{i=1}^{n}\sum_{j=1}^{m_i} \log(1 + e^{-y_{ij}*F(q_i,a_i,r_{ij})})$, where $y_{ij}$ is the label for $r_{ij}$.

\noindent
\softmax: We will discuss in \S \ref{sec:dataset} that annotators are expected to miss a few good responses since good and bad answers are often very similar (may only differ by a single preposition or pronoun). Therefore, we explore a ranking objective that calculates errors based on the margin with which incorrect responses are ranked above correct ones \citep{collins-koo-2005-discriminative}.
Without loss of generality, we assume $r_{i1}$ to be better than all other responses for $(q_i, a_i)$. Since the model $F()$ should rank $r_{i1}$ higher than all other responses, we use the margin error $M_{ij}(F) = F(q_i, a_i, r_{i1}) - F(q_i, a_i, r_{ij})$ to define the {\softmax} loss 
as $\sum_{i=1}^{n} \log{\left ( 1+\sum_{j=2}^{m_i}e^{-M_{ij}(F)} \right )}$. 

\par We experiment with the following feature based and neural models with the two loss functions:

\noindent
\textbf{Language Model Baseline}: The responses are ranked using the normalized probabilities from a 3-gram LM trained on the Gigaword corpus with modified Kneser-Ney smoothing.\footnote{\href{http://www.keithv.com/software/giga/}{http://www.keithv.com/software/giga/}} The response with the highest score is classified as 1 and others as 0.

\noindent
\textbf{Linear Model}: A linear classifier using features inspired by \citet{heilman-smith-2010-good} and \citet{wan2006using}, who have implemented similar linear models for other sentence pair classification tasks. Specifically, we use the following features:
\begin{itemize}[leftmargin=*]
    \itemsep-1mm 
    \item Length \textbf{(Features 1-3)}: word length of question $q_i$, answer-phrase $a_i$, and response $r_{ij}$
    \item WH-word \textbf{(Features 4-12)}: [0-1 feat.] \textit{what}, \textit{who}, \textit{whom}, \textit{whose}, \textit{when}, \textit{where}, \textit{which}, \textit{why} or \textit{how} is present in the $q_i$
    \item Negation \textbf{(Features 13)}: [0-1 feat.]  \textit{no}, \textit{not} or \textit{none} is present in the $q_i$
    \item N-gram LM \textbf{(Features 14-21)}: 2, 3-gram normalized probability and perplexity of $q_i$ and $r_{ij}$
    \item Grammar \textbf{(Features 22-93)}: node counts of $q_i$ and $r_{ij}$ syntactic parse trees
    \item Word overlap \textbf{(Features 94-96)}: three features based on fraction of word overlap between $q_i$ and $r_{ij}$. $precision = \frac{overlap(q_i, r_{ij})}{|q_i|}$, $recall = \frac{overlap(q_i, r_{ij})}{|r_{ij}|}$ and their harmonic mean
\end{itemize}

\noindent
\textbf{Decomposable Attention}: We use the sentence pair classifier from \citep{parikh-etal-2016-decomposable}, referred as the \textbf{DA} model. It finds attention based word-alignment of the input pair (premise and hypothesis, in our case question $q_i$ and response $r_{ij}$) and aggregates it using feedforward networks.
Apart from standard vector embeddings, we also experiment with contextualized {\elmo} \citep{peters-etal-2018-deep} embedding with the \textbf{DA} model using the version implemented in AllenNLP \citep{Gardner2017AllenNLP}.

\noindent
\textbf{BERT}: Lastly, we use the BERT-Base, Uncased model \citep{devlin-etal-2019-bert} for sentence pair classification. The model takes question $q_i$ and response $r_{ij}$ separated by the special token \texttt{[SEP]} and predicts if the response is suitable or unsuitable.

In some cases, the number of responses generated by the STs for a question could be as high as 5000+. Therefore, when training the \textbf{DA} model with pre-trained contextualized embeddings such as ELMo or the \textbf{BERT} model in the {\softmax} loss setting, backpropagation requires computing and storing hidden states for 5000+ different responses. 
To mitigate this issue, we use \textit{strided negative-sampling}. While training, we first separate all the suitable responses 
from all the remaining unsuitable responses. We then divide all the responses for $q_i$ into smaller batches of $K$ or fewer responses. Each batch comprises one suitable response (randomly chosen) and $K-1$ sampled from the unsuitable responses. To ensure that all unsuitable responses are used at least once during the training, we shuffle them and then create smaller batches by taking strides of $K-1$ size. We use $K=150$ for \textbf{DA}+{\elmo} and $K=50$ for \textbf{BERT} when training with the {\softmax} loss. At test time, we compute logits on the CPU and normalize across all responses.

\subsection{Training Data for Response Classification}
\label{sec:dataset}

In this section, we describe the details of the training, validation and testing data used to develop the {\em best response classifier} models. To create the supervised data, we choose a sample from the \textit{train-set} of the {\squadtwo} dataset \citep{rajpurkar-etal-2018-know}. {\squadtwo} contains human-generated questions and answer spans selected from Wikipedia paragraphs. Before sampling, we remove all the QA pairs which had answer spans $>5$ words as they tend to be non-factoid questions and complete sentences in themselves (typically \quotes{why} and \quotes{how} questions). We also filter out questions that cannot be handled by the parser ($\sim 20\%$ of them had obvious parser errors). After these filtering, we take a sample of 3000 questions and generate their list of responses using STs (1,561,012 total responses).

\par Next, we developed an annotation task on Amazon Mechanical Turk to select the best responses for the questions. For each question, we ask the annotators to select a response from the list of responses that correctly answers the question, sounds natural, and seems human-like. Since the list of responses for some questions is as long as 5000+, the annotators can't review all of them before selecting the best one. Hence, we implement a search feature within the responses list such that annotators can type in a partial response in the search box to narrow down the options before selection. To make their job easier, we also sorted responses by length. This encouraged annotators to select relatively short responses which we found to be beneficial, as one would prefer an automatic QA system to be terse. To verify that the annotators didn't cheat this annotation design by selecting the first/shortest option, we also test a \textbf{Shortest Response Baseline} as another baseline response classifier model, where first/shortest response in the list is selected as suitable.
\par Each question is assigned 5 annotators. Therefore, there can be at most 5 unique annotated responses for each question. This decreases the recall of the gold truth data (since there can be more than 5 good ways of correctly responding to a question). On the other hand, bad annotators may choose a unique yet suboptimal/incorrect response, which decreases the precision of the gold truth.

\par After annotating the 3000 questions from {\squadtwo} sample, we randomly split the data into 2000 train, 300 validation, and 700 test questions. We refer to this as the SQuAD Gold annotated (\textbf{SG}) data. To increase \textbf{SG} training data precision, we assign label 1 only to responses that are marked as best by at least two different annotators. Due to this hard constraint, 244 questions from the training data are removed (i.e. the 5 annotators marked 5 unique responses). On the other hand, to increase the recall of the \textbf{SG} test and validation sets, we retain all annotations.\footnote{We found that some bad annotators had a high affinity of choosing the first (or the shortest) response when it was not the best choice in the list. To reduce such annotation errors we add another constraint that the shortest response should be selected by at least 2 different annotators.} We assign label 0 to all remaining responses (even if some of them are plausible). The resulting \textbf{SG} data split is summarized in Table~\ref{tab:SQuAD_sample_data}. 
\begin{table}[t]
\fontsize{10}{11}\selectfont
\centering
\begin{tabular}{|l|l|l|l|}

\hline
      & $\#q/\#a$  & \cmark $\#r$ & \xmark $\#r$ \\ \hhline{|=|=|=|=|}
Train & 1756 & 2028    & 796174  \\ \hline
Val   & 300  & 791     & 172135  \\ \hline
Test  & 700  & 1833     & 182963  \\ \hline
\end{tabular}
\caption{Statistics of the \textbf{SG} training, validation, and test sets curated from the {\squadtwo} \textit{training} data. $q$ and $a$ denotes the question and answer from the {\squadtwo} sample and $r$ denotes the responses generated by the STs. $\#q$ means \quotes{number of questions}. \cmark $\#r$ and \xmark $\#r$ denotes the number of responses which are labeled 1 and 0 respectively after the human annotation process.}
\label{tab:SQuAD_sample_data}
\end{table}

\par Every response may be marked by zero or more annotators. When at least two annotators select the same response from the list we consider it as a \textit{match}. To compute the annotator agreement score, we divide the number of matches with total number of annotations by each annotator. Using this formula we find average annotator agreement to be 0.665, where each annotator's agreement score is weighted by their number of annotated questions.

\subsection{Evaluation of Response Classification}
As previously mentioned in \S \ref{sec:dataset}, the \textbf{SG} data doesn't contain all true positives since one cannot exhaustively find and annotate all the good responses when the response list is very long.  Additionally, there is a large class imbalance between good and bad responses, making standard evaluation metrics such as precision, recall, F1 score and accuracy potentially misleading. To gather additional insight regarding how well the model ranks correct responses over incorrect ones, we calculate Precision@1 (P@1),\footnote{P@1 is the \% of times the correct response is ranked first} Max. F1,\footnote{Max. F1 is the maximum F1 the model can achieve by choosing the optimal threshold in the PR curve} and Area Under the Precision-Recall Curve (PR-AUC). We train all classifier models on the \textbf{SG} training set and evaluate them on \textbf{SG} test data. The resulting evaluation is presented in Table \ref{tab:auto_eval}.
\begin{table}[t]
\centering
\fontsize{10}{11}\selectfont
\begin{tabular}{|l|l|l|l|l|}
\hline
Classifier & Loss & P@1 & Max-F1 & PR-AUC  \\ \hhline{|=|=|=|=|=|}
ShortResp & - & 0.324 & 0.189 & - \\ \hline
LangModel & - & 0.058  &  0.012 & - \\ \hline
Linear   & Log. & 0.680 & 0.159 & 0.070 \\ \hline
Linear   & Soft. & 0.640 & 0.387 & 0.344 \\ \hline
DA        & Log. & 0.467 & 0.151 & 0.066 \\ \hline
DA+ELMo  & Log. & 0.694      &  0.354 & 0.301 \\ \hline
DA          & Soft. & 0.503 &  0.383 & 0.297 \\ \hline
DA+ELMo   & Soft. & 0.716 & 0.456 & 0.427 \\ \hline
BERT   & Log. & 0.816 & 0.490 & \textbf{0.465} \\ \hline
BERT   & Soft. & \textbf{0.833} & \textbf{0.526} & 0.435 \\ \hline
\end{tabular}
\caption{Best response classifier results on \textbf{SG} test data. \quotes{ShortResp} stands for Shortest Response baseline, \quotes{LangModel} stands for Language Model baseline, \quotes{Linear} stands for Linear model. \quotes{Log.} and \quotes{Soft.} in Loss column stands for Logistic and Softmax loss respectively. DA refers to Decomposable Attention model \citep{parikh-etal-2016-decomposable}. 
\quotes{+{\elmo}} refers to adding pre-trained {\elmo} embeddings to DA model.
}
\label{tab:auto_eval}
\end{table}

The results show that the shortest response baseline (ShortResp) performs worse than the ML models (0.14 to 0.51 absolute P@1 difference depending on the model). This verifies that annotation is not dominated by presentation bias where annotators are just selecting the shortest (first in the list) response for each question. The language model baseline (LangModel) performs even worse (0.41 to 0.78 absolute difference), demonstrating that this task is unlikely to have a trivial solution. The feature-based linear model shows good performance when trained with {\softmax} loss beating many of the neural models in terms of PR-AUC and Max-F1. By inspecting the weight vector, we find that grammar features, specifically the number of prepositions, determiners, and \quotes{to}s in the response, are the features with the highest weights. This probably implies that the most important challenge in this task is finding the right prepositions and determiners in the response. Other important features are the response length and the response's 3-gram LM probabilities. The ostensible limitation of feature-based models is failing to recognize correct pronouns for unfamiliar named entities in the questions.

Due to the small size of \textbf{SG} train set, the vanilla Decomposable Attention (\textbf{DA}) model is unable to learn good representations on its own and accordingly, performs worse than the linear feature-based model. The addition of {\elmo} embeddings appears to help to cope with this. We find that the \textbf{DA} model with {\elmo} embeddings is better able to predict the right pronouns for the named entities, presumably due to pre-trained representations.
The best neural model in terms of P@1 is the \textbf{BERT} model fine-tuned with the {\softmax} loss (last row of Table \ref{tab:auto_eval}).

\section{Data-Augmentation and Generation}
\label{sec:aug_and_generate}
{\SeqToSeq} models are very effective in generation tasks. However, our 2028 labeled question and response pairs from the \textbf{SG} train set (Table \ref{tab:SQuAD_sample_data}) are insufficient for training these large neural models. On the other hand, creating a new large-scale dataset that supports fluent answer generation by crowdsourcing is inefficient and expensive. Therefore, we augment {\squadtwo} with responses from the STs+\textbf{BERT} classifier (Table \ref{tab:auto_eval}) to create a synthetic training dataset for {\SeqToSeq} models. We take all the QA pairs from the {\squadtwo} \textit{train-set} which can be handled by the question parser and STs, and rank their candidate responses using the \textbf{BERT} response classifier probabilities trained with {\softmax} loss (i.e. ranking loss \citep{collins-koo-2005-discriminative}). Therefore, for each question we select the top ranked responses\footnote{at most three responses per question} by setting a threshold on the probabilities obtained from the \textbf{BERT} model. We refer to the resulting dataset as SQuAD-Synthetic (\textbf{SS}) consisting of 59,738 $\langle q,a,r\rangle$ instances. 
\par 
To increase the size of \textbf{SS} training data, we take the QA pairs from Natural Questions \citep{kwiatkowski-etal-2019-natural} and HarvestingQA\footnote{HarvestingQA is a QA dataset containing 1M QA pairs generated over 10,000 top-ranking Wikipedia articles. This dataset is noisy as the questions are automatically generated using an LSTM based encoder-decoder model (which makes use of coreference information) and the answers are extracted using a candidate answer extraction module.} \citep{du-cardie-2018-harvesting} and add $\langle q,a,r\rangle$ instances using the same STs+\textbf{BERT} classifier technique. These new pairs combined with \textbf{SS} result in a dataset of 1,051,938 $\langle q,a,r\rangle$ instances, referred to as the  \textbf{SS+} dataset.

\subsection{PGN, D-GPT, Variants and Baselines}
\label{subsec:models}
\par Using the resulting \textbf{SS} and \textbf{SS+} datasets, we train Pointer generator networks (PGN) \citep{see-etal-2017-get}, DialoGPT (D-GPT) \citep{zhang2019dialogpt} and their variants to produce a fluent answer response generator. The input to the generation model is the question and the answer phrase $\langle q,a\rangle$ and the response $r$ is the corresponding generation target.

\noindent
 \textbf{{\pgn}}: PGNs are widely used {\SeqToSeq} models equipped with a copy-attention mechanism capable of copying any word from the input directly into the generated output, making them well equipped to handle rare words and named entities present in questions and answer phrases. We train a 2-layer stacked bi-LSTM PGN using the OpenNMT toolkit \citep{opennmt} on the \textbf{SS} and \textbf{SS+} data. We additionally explore PGNs with pre-training information by initializing the embedding layer with {\glove} vectors \citep{pennington-etal-2014-glove} and pre-training it with $\langle q,r\rangle$ pairs from the questions-only subset of the OpenSubtitles corpus\footnote{\href{http://forum.opennmt.net/t/english-chatbot-model-with-opennmt/184}{http://forum.opennmt.net/t/english-chatbot-model-with-opennmt/184}} \citep{tiedemann2009news}. This corpus contains about 14M question-response pairs in the training set and 10K pairs in the validation set. We name the pre-trained {\pgn} model as {\pgnp}. We also fine-tune {\pgnp} on the \textbf{SS} and \textbf{SS+} data to generate two additional variants.

\noindent
 \textbf{{\dgpt}}: DialoGPT (i.e. dialogue generative pre-trained transformer)~\citep{zhang2019dialogpt} is a recently released large tunable automatic conversation model trained on 147M Reddit conversation-like exchanges using the GPT-2 model architecture~\citep{radford2019language}. We fine-tune {\dgpt} on our task using the \textbf{SS} and \textbf{SS+} datasets. For comparison we also train {\gpt} on our datasets from scratch (i.e. without any pre-training). Finally, to assess the impact of pre-training datasets, we pre-train the {\gpt} on the 14M questions from questions-only subset of the OpenSubtitles data (similar to the {\pgnp} model) to get {\gptp} model. The {\gptp} is later fine-tuned on the \textbf{SS} and \textbf{SS+} datasets to get two corresponding variants.

\begin{table*}[ht]
\centering
\fontsize{10}{11}\selectfont
\begin{tabular}{|l|c|c|c|c|c|c||c||c|}
\cline{1-8} 
\multirow{4}{*}{Model}& \multirow{4}{*}{Data}      &  \multirow{4}{*}{PPL} & \texttt{a} & \texttt{b} & \texttt{c} & \texttt{d} & \texttt{e} \\ \cline{4-9}
            &  & & \xmark & \cmark & \xmark & \cmark & \cmark & correct answer \\ \cline{4-9}
            & & & \xmark & \xmark & \cmark & \cmark & \cmark & complete-sentence  \\\cline{4-9}
            &  & & -      & -      & -      & \xmark & \cmark & grammaticality \\ \hhline{|=|=|=|=|=|=|=||=||-|}
            
{{\coqa} B.}& -   & - &  13.80   &       82.20   &       1.20    &       0.60    &       2.20  \\ \cline{1-8}
{{\quac} B.} & -     & - &  5.20    &       3.80    &       46.40   &       2.80    &       41.80 \\ \cline{1-8}
{STs+BERT B.} & - & - &  0.00    &       18.20   &       0.20    &       13.80   &       67.80  \\\hhline{|=|=|=|=|=|=|=||=||~}
{\pgn} & SS           & 6.60 & 1.00    &       7.00    &       9.00    &       16.20   &       66.80 \\ \cline{1-8}
{\pgn} & SS+         & 3.83 &  1.00    &       3.00    &       8.40    &       17.60   &       70.00 \\ \cline{1-8}
{\pgnp} & SS    & 4.34 &   0.20    &       4.60    &       9.80    &       17.40   &       68.00  \\ \cline{1-8}
{\pgnp} & SS+       & 3.31 &  0.40    &       4.80    &       9.00    &       16.20   &       69.60 \\ \cline{1-8}
{\gpt} & SS           & 4.69 &  1.00    &       5.00    &       13.20   &       18.60   &       62.20 \\ \cline{1-8}
{\gpt} & SS+         & 2.70 &   0.80    &       4.20    &       8.20    &       16.80   &       70.00 \\ \cline{1-8}
{\gptp} & SS    & 3.23 &   0.40    &       2.80    &       8.20    &       19.00   &       69.60  \\ \cline{1-8}
{\gptp} & SS+       & 2.74 &   0.80    &       2.40    &       7.80    &       17.00   &       72.00 \\ \cline{1-8} 
{\dgpt} & SS    & 2.20 & 0.40    &       2.40    &       8.60    &       13.00   &       75.60  \\ \cline{1-8}
{\dgpt} & SS+       & 2.06 &  0.40    &       2.60    &       7.80    &       13.20   &       \textbf{76.00} \\\hhline{|=|=|=|=|=|=|=||=||~}
{\dgpt} (o) & SS+       & 2.06 &   0.00    &       3.00    &       0.00    &       13.80   &       \textbf{83.20} \\ \cline{1-8} 
\end{tabular}
\caption{Human evaluation results of all the models and baselines on sample of \textit{SQuAD-dev-test}. In the first three rows B. stands for baseline. In the last row "(o)" stands for oracle. In Column 3 PPL stands for validation perplexity. All the values are percentage (out of 100) of responses from each model that belong to specific option(\texttt{a} to \texttt{e}) selected by annotators.}
\label{tab:squad_dev_test_results}
\vspace{-5mm}
\end{table*}

\noindent
 \textbf{{\coqa} Baseline}: \textbf{Co}nversational \textbf{Q}uestion \textbf{A}nswering (CoQA) \citep{reddy-etal-2019-coqa} is a large-scale ConvQA dataset aimed at creating models which can answer the questions posed in a conversational setting. Since we are generating conversational responses for QA systems, it is sensible to compare against such ConvQA systems. We pick one of the best performing BERT-based CoQA model from the SMRCToolkit~\citep{2019arXiv190311848W} as a baseline.\footnote{one of the top performing model with available code.} We refer to this model as the \textbf{\coqa} baseline.

\noindent
 \textbf{{\quac} Baseline}: \textbf{Qu}estion \textbf{A}nswering in \textbf{C}ontext is another ConvQA dataset. We use the modified version of BiDAF model presented in \citep{choi-etal-2018-quac} as a second baseline. Instead of a {\SeqToSeq} generation, it selects spans from passage which acts as responses. We use the version of this model implemented in AllenNLP \citep{Gardner2017AllenNLP} and refer to this model as the \textbf{\quac} baseline.

\noindent
 \textbf{STs+BERT Baseline}: We also compare our generation models with the technique that created the \textbf{SS} and \textbf{SS+} training datasets (i.e. the responses generated by STs ranked with the \textbf{BERT} response classifier).
 
We validate all the {\SeqToSeq} models on the human annotated \textbf{SG} data (Table \ref{tab:SQuAD_sample_data}).

\subsection{Evaluation on the SQuAD 2.0 Dev Set}
\label{sec:squad_experiments}
To have a fair and unbiased comparison, we create a new 500 question sample from the SQuAD 2.0 \textit{dev} set (\textit{SQuAD-dev-test}) which is unseen for all the models and baselines. This sample contains $\sim 20\%$ of the questions that cannot be handled by the STs (parser errors). For such questions, we default to outputting the answer-phrase as the response for the \textbf{STs+BERT} baseline. For the \textbf{\coqa} baseline and the \textbf{\quac} baseline, we run their models on passages (corresponding to the questions) from \textit{SQuAD-dev-test} to get their responses.

To demonstrate that our models too can operate in a fully automated setting like the \textbf{\coqa} baseline and the \textbf{\quac} baseline, we generate their responses using the answer spans selected by a BERT-based SQuAD model (instead of the gold answer span from the \textit{SQuAD-dev-test}). 

For automatic evaluation we compute validation perplexity of all {\SeqToSeq} generation models on \textbf{SG} data ($3^{rd}$ column in Table \ref{tab:squad_dev_test_results}). However, validation perplexity is a weak evaluator of generation models. Also, due to the lack of human-generated references in  \textit{SQuAD-dev-test}, we cannot use other typical generation based automatic metrics. Therefore, we use Amazon Mechanical Turk to do human evaluation. Each response is judged by 5 annotators. We ask the annotators to identify if the response is conversational and answers the question correctly. While outputting answer-phrase to all questions is trivially correct, this style of response generation seems robotic and unnatural in a prolonged conversation. Therefore, we also ask the annotators to judge if the response is a complete-sentence (e.g. \quotes{it is in Indiana}) and not a sentence-fragment (e.g. \quotes{Indiana}). For each question and response pair, we show the annotators five options based on the three properties (correctness, grammaticality, and complete-sentence). These five options (\texttt{a} to \texttt{e}) are shown in the Table \ref{tab:squad_dev_test_results} header. The best response is a complete-sentence which is grammatical and answers the question correctly (i.e. option \texttt{e}). Other options give us more insights into different models' behavior. For each response, we assign the majority option selected by the annotators and aggregate their judgments into buckets. We present this evaluation in Table \ref{tab:squad_dev_test_results}. 

\par We compute the inter-annotator agreement by calculating Cohen's kappa \citep{cohen1960coefficient} between individual annotator's assignments and the aggregated majority options. The average Cohen's kappa (weighted by the number of annotations for every annotator) is 0.736 (i.e. substantial agreement).

The results reveal that \textbf{\coqa} baseline does the worst in terms of option \texttt{e}. The main reason for that is because most of the responses generated from this baseline are exact answer spans. Therefore, we observe that it does very well in option \texttt{b} (i.e. correct answer but not a complete-sentence). The \textbf{\quac} baseline can correctly select span-based informative response $\sim 42\%$ of the time. Other times, however, it often selects a span from the passage which is related to the topic but doesn't contain the correct answer i.e. (option \texttt{c}). Another problem with this baseline is that it is restricted by the input passage and many not always be able to find a valid span that answers the questions. Our \textbf{STs+BERT} baseline does better in terms of option \texttt{e} compared to the other baselines but it is limited by the STs and the parser errors. As mentioned earlier, $\sim 20\%$ of the time this baseline directly copies the answer-phrase in the response which explains the high percentage of option \texttt{b}.
\par Almost all models perform better when trained with \textbf{SS+} data showing that the additional data from Natural Questions and HarvestingQA is helping. Except for the {\pgn} model trained on \textbf{SS} data, all other variants perform better than \textbf{STs+BERT} baseline in terms of option \texttt{e}. The {\gpt} model trained on \textbf{SS} data from scratch does not perform very well because of the small size of training data. The pretraining with OpenSubtitiles questions boosts its performance (option \texttt{e} \% for {\gptp} model variants $>$ option \texttt{e} \% for {\gpt} model variants). The best model however is {\dgpt} when finetuned with \textbf{SS+} dataset. While retaining the correct answer, it makes less grammatical errors (lower \% in option \texttt{c} and \texttt{d} compared to other models). Furthermore with oracle answers it performs even better (last row in Table \ref{tab:squad_dev_test_results}). This shows that {\dgpt} can generate better quality responses with accurate answers. We provide some sample responses from different models in Appendix \ref{sec:sample_responses}.

\begin{table}[t]
\centering
\fontsize{10}{11}\selectfont
\begin{tabular}{|l|c|c|c|c||c||}
\hline 
Model    & \texttt{a} & \texttt{b} & \texttt{c} & \texttt{d} & \texttt{e} \\ \hhline{|=|=|=|=|=||=||}
{{\coqa} B.}  &   12.0   &       78.0   &       5.0    &       2.0    &       3.0  \\ \hline
{\dgpt}  &    2.0    &       5.0    &       16.0   &       20.0   &       \textbf{57.0} \\\hhline{|=|=|=|=|=||=||}
{\dgpt} (o)      &   0.0    &       7.0    &       0.0    &       16.0   &       \textbf{77.0} \\ \hline
\end{tabular}
\caption{Human evaluation results of \textbf{\dgpt} model (trained on \textbf{SS+} dataset) vs \textbf{\coqa} model on sample of 100 question answers from filtered {\coqa} dev set. (o) stands for oracle answers. Options \texttt{a} to \texttt{e} are explained in Table \ref{tab:squad_dev_test_results} header. }
\label{tab:coqa_eval}
\end{table}

\subsection{Evaluation on CoQA}
\label{sec:cross_domain_experiments}
In this section, we test our model's ability to generate conversational answers on the {\coqa} dev set, using \textbf{\coqa} baseline's predicted answers. The {\coqa} dataset consists of passages from seven different domains (out of which one is Wikipedia) and conversational questions and answers on those passages. Due to the conversational nature of this dataset, some of the questions are one word ($\sim 3.1\%$), like \quotes{what?}, \quotes{why?} etc. Such questions are out-of-domain for our models as they require the entire context over multiple turns of the conversation to develop their response. Other out-of-domain questions include unanswerable ($\sim 0.8\%$) and yes/no ($\sim 18.4\%$) questions. We also don't consider questions with answers $> 5$ words ($\sim 11.6\%$) as they are typically non-factoid questions. We take a random sample of 100 from the remaining questions. This sample contains questions from a diverse set of domains outside of the Wikipedia (on which our models are trained). This includes questions taken from the middle of a conversation (for example, \quotes{who did they meet ?}) which are unfamiliar for our models. We perform a human evaluation similar to \S \ref{sec:squad_experiments} on this sample. We compare \textbf{\coqa} against \textbf{\dgpt} trained on the \textbf{SS+} dataset (with \textbf{\coqa}'s predictions input as answer-phrases). The results are shown in Table \ref{tab:coqa_eval}.

\par This evaluation reveals that the \textbf{\dgpt} model is able to successfully convert the \textbf{\coqa} answer spans into conversational responses 57\% of the time (option \texttt{e}). \textbf{\dgpt} gets the wrong answer 18\% of the time (option \texttt{a} and \texttt{c}), because the input answer predicted by the \textbf{CoQA} baseline is also incorrect 17\% of the time. However with oracle answers, it is able to generate correct responses 77\% of the times (option \texttt{e}). The weighted average Cohen's kappa \citep{cohen1960coefficient} score for all annotators in this evaluation is 0.750 (substantial agreement). This result demonstrates ability of our model to generalize over different domains and generate good conversational responses for questions when provided with correct answer spans.

\section{Related Work}
\textbf{Question Generation} (QG) is a well studied problem in the NLP community with many machine learning based solutions \cite{rus2010first, heilman-smith-2010-good, yao2012semantics, labutov-etal-2015-deep, serban-etal-2016-generating, reddy-etal-2017-generating, du-etal-2017-learning, du-cardie-2017-identifying, du-cardie-2018-harvesting}. In comparison, our work explores the opposite direction, i.e. (generating conversational humanlike answers given a question). \citet{fu-feng-2018-natural} also try to solve fluent answer response generation task but in a restricted setting of movie related questions with 115 question patterns. In contrast, our generation models can deal with human generated questions from any domain.
Most closely related to our approach,
\citet{demszky2018transforming} presented a rule-based method, and a neural sequence model to convert QA pairs into a single declarative answer sentence.  Their main goal was to automatically generate natural language inference datasets from existing QA datasets, by pairing a generated declarative sentence with an associated passage of text.  In contrast, we used an approach based on over-generating and ranking candidate responses, which were then used to fine-tune pre-trained dialogue models, such as DialoGPT.
\par \textbf{Learning to Rank} formulations for answer selection in QA systems is common practice, most frequently relying on {\em pointwise} ranking models~\cite{severyn2015learning, garg2019tanda}. Our use of discriminative re-ranking~\citep{collins-koo-2005-discriminative} with softmax loss is closer to learning a {\em pairwise} ranking by maximizing the multiclass margin between correct and incorrect answers~\citep{joachims2002optimizing, burges2005learning, koppel2019pairwise}. 
This is an important distinction from TREC-style answer selection as our ST-generated candidate responses have lower semantic, syntactic, and lexical variance, making pointwise methods less effective. 

\textbf{Question Answering} Using crowd-sourcing methods to create QA datasets \cite{rajpurkar-etal-2016-squad, bajaj2016ms, rajpurkar-etal-2018-know}, conversational datasets~\cite{dinan2018wizard}, and ConvQA datasets \cite{choi-etal-2018-quac, reddy-etal-2019-coqa, elgohary-etal-2018-dataset, saha2018complex} has largely driven recent methodological advances. However, models trained on these ConvQA datasets typically select exact answer spans instead of generating them \cite{yatskar-2019-qualitative}.
Instead of creating another crowd-sourced dataset for our task, we augment existing QA datasets to include such conversational answer responses using the STs + BERT trained with softmax loss. 

\section{Conclusion}

In this work, we study the problem of generating fluent QA responses in the context of building fluent conversational agents. To this end, we propose an over-generate and rank data augmentation procedure based on {\lgr} and a best response classifier. This method is used to modify the {\squadtwo} dataset such that it includes conversational answers, which is used to train {\SeqToSeq} based generation models. Human evaluations on \textit{SQuAD-dev-test} show that our models generate significantly better conversational responses compared to the baseline {\coqa} and {\quac} models. Furthermore, the {\dgpt} model with oracle answers is able to generate conversational responses on the {\coqa} dev set 77 \% of the time showcasing the model's scalability.


\subsubsection*{Acknowledgments}
We would like to thank Greg Durrett and the anonymous reviewers for valuable feedback on an earlier draft of this paper.  This material is based in part on research sponsored by the NSF (IIS-1845670), ODNI and IARPA via the BETTER program (2019-19051600004) DARPA via the ARO (W911NF-17-C-0095) in addition to an Amazon Research Award. The views and conclusions contained herein are those of the authors and should not be interpreted as necessarily representing the official policies, either expressed or implied, of NSF, ODNI, ARO, IARPA, DARPA or the U.S. Government.

\bibliography{acl2020}
\bibliographystyle{acl_natbib}

\appendix

\clearpage
\onecolumn
\section{Sample responses from different models}
\label{sec:sample_responses}

\fontsize{10}{11}\selectfont
\begin{longtable}{|l|p{258px}|l|l|l|l|}
\hline
\multicolumn{6}{|p{\linewidth}|}{Sample of responses from different models on \textit{SQuAD-dev-test} set \S \ref{sec:squad_experiments}. "\textbf{Q}:" is the Question, "\textbf{R}:" is the Response, "B." stands for baseline and "(o)" stands for oracle answer spans.}\\
\hline
Model  & \textbf{Q}:what fixed set of factors determine the actions of a deterministic turing machine & \rotatebox[origin=c]{270}{correctness} & \rotatebox[origin=c]{270}{complete-sentence} & \rotatebox[origin=c]{270}{grammaticality} & \rotatebox[origin=c]{270}{majority option}  \\ \hline 
{\coqa} B. & \textbf{R}: fixed set of rules & \cmark & \xmark & - & b \\ \hline
{\quac} B. & \textbf{R}: a deterministic turing machine is the most basic turing machine , which uses a fixed set of rules to determine its future actions & \cmark & \cmark & \cmark & e \\ \hline
STs+BERT B. & \textbf{R}: rules determine the actions & \cmark & \cmark & \cmark & e \\ \hline
{\pgnp} with SS+ & \textbf{R}: fixed set of rules determine the actions & \cmark & \cmark & \cmark & e \\ \hline
{\gpt} with SS+ & \textbf{R}: fixed set of rules fixed set of factors determine the actions & \cmark & \cmark & \xmark & d \\ \hline
{\gptp} with SS+ & \textbf{R}: fixed set of rules determine the actions & \cmark & \cmark & \cmark & e \\ \hline
{\dgpt} with SS+ & \textbf{R}: fixed set of rules determine the actions & \cmark & \cmark & \cmark & e \\ \hline
{\dgpt} with SS+ (o) & \textbf{R}: rules determine the actions & \cmark & \cmark & \cmark & e \\ \hline
Model  & \textbf{Q}:why was polonia relegated from the country 's top flight in 2013 ? & \rotatebox[origin=c]{270}{correctness} & \rotatebox[origin=c]{270}{complete-sentence} & \rotatebox[origin=c]{270}{grammaticality} & \rotatebox[origin=c]{270}{majority option}  \\ \hline 
{\coqa} B. & \textbf{R}: because of their disastrous financial situation & \cmark & \xmark & - & b \\ \hline
{\quac} B. & \textbf{R}: polonia was relegated from the country 's top flight in 2013 because of their disastrous financial situation & \cmark & \cmark & \cmark & e \\ \hline
LGRs+BERT B. & \textbf{R}: polonia was relegated from the country 's top flight because the disastrous financial situation & \cmark & \cmark & \xmark & d \\ \hline
{\pgnp} with SS+ & \textbf{R}: polonia was relegated disastrous financial situation & \cmark & \cmark & \xmark & d \\ \hline
{\gpt} with SS+ & \textbf{R}: polonia was relegated disastrous financial situation & \cmark & \cmark & \xmark & d \\ \hline
{\gptp} with SS+ & \textbf{R}: polonia was relegated disastrous financial situation & \cmark & \cmark & \xmark & d \\ \hline
{\dgpt} with SS+ & \textbf{R}: polonia was relegated from the country's top flight for disastrous financial situation & \cmark & \cmark & \cmark & e \\ \hline
{\dgpt} with SS+ (o) & \textbf{R}: polonia was relegated from the country's top flight for disastrous financial situation & \cmark & \cmark & \cmark & e \\ \hline
Model  & \textbf{Q}:how was sadat rewarded by the islamists for his attempts to bring egypt into modern times and civilization ? & \rotatebox[origin=c]{270}{correctness} & \rotatebox[origin=c]{270}{complete-sentence} & \rotatebox[origin=c]{270}{grammaticality} & \rotatebox[origin=c]{270}{majority option}  \\ \hline 
{\coqa} B. & \textbf{R}: released Islamists from prison & \xmark & \xmark & - & a \\ \hline
{\quac} B. & \textbf{R}: egyptian president anwar sadat – whose policies included opening egypt to western investment ( infitah ) ; transferring egypt 's allegiance from the soviet union to the united states & \xmark & \cmark & - & c \\ \hline
LGRs+BERT B. & \textbf{R}: sadat was rewarded by the islamists for his attempts to bring egypt assassinated & \cmark & \cmark & \xmark & d \\ \hline
{\pgnp} with SS+ & \textbf{R}: sadat was rewarded piety and charity & \xmark & \cmark & - & c \\ \hline
{\gpt} with SS+ & \textbf{R}: it was rewarded piety and charity & \xmark & \cmark & - & c \\ \hline
{\gptp} with SS+ & \textbf{R}: sadat was rewarded by piety and charity & \xmark & \cmark & - & c \\ \hline
{\dgpt} with SS+ & \textbf{R}: sadat was rewarded piety and charity & \xmark & \cmark & - & c \\ \hline
{\dgpt} with SS+ (o) & \textbf{R}: sadat was assassinated & \cmark & \cmark & \cmark & e \\ \hline
Model  & \textbf{Q}:from what type of materials must liquid oxygen be separated ? & \rotatebox[origin=c]{270}{correctness} & \rotatebox[origin=c]{270}{complete-sentence} & \rotatebox[origin=c]{270}{grammaticality} & \rotatebox[origin=c]{270}{majority option}  \\ \hline 
{\coqa} B. & \textbf{R}: combustible materials & \cmark & \xmark & - & b \\ \hline
{\quac} B. & \textbf{R}: both liquid and solid o \ n2 are clear substances with a light sky-blue color caused by absorption in the red & \xmark & \cmark & - & c \\ \hline
STs+BERT B. & \textbf{R}: combustible materials & \cmark & \xmark & - & b \\ \hline
{\pgnp} with SS+ & \textbf{R}: it must be separated from combustible & \cmark & \cmark & \cmark & e \\ \hline
{\gpt} with SS+ & \textbf{R}: it must be separated from combustible & \cmark & \cmark & \cmark & e \\ \hline
{\gptp} with SS+ & \textbf{R}: they must be separated from combustible & \cmark & \cmark & \cmark & e \\ \hline
{\dgpt} with SS+ & \textbf{R}: liquid oxygen must be separated from combustible & \cmark & \cmark & \cmark & e \\ \hline
{\dgpt} with SS+ (o) & \textbf{R}: liquid oxygen must be separated from combustible materials & \cmark & \cmark & \cmark & e \\ \hline
Model  & \textbf{Q}:on what type of exponentiation does the diffie – hellman key exchange depend on ? & \rotatebox[origin=c]{270}{correctness} & \rotatebox[origin=c]{270}{complete-sentence} & \rotatebox[origin=c]{270}{grammaticality} & \rotatebox[origin=c]{270}{majority option}  \\ \hline 
{\coqa} B. & \textbf{R}: modular exponentiation & \cmark & \xmark & - & b \\ \hline
{\quac} B. & \textbf{R}: several public-key cryptography algorithms , such as rsa and the diffie – hellman key exchange & \xmark & \xmark & - & a \\ \hline
STs+BERT B. & \textbf{R}: modular exponentiation & \cmark & \xmark & - & b \\ \hline
{\pgnp} with SS+ & \textbf{R}: it depends on modular & \cmark & \cmark & \cmark & e \\ \hline
{\gpt} with SS+ & \textbf{R}: it depends on modular & \cmark & \cmark & \cmark & e \\ \hline
{\gptp} with SS+ & \textbf{R}: it depends on the modular & \cmark & \cmark & \cmark & e \\ \hline
{\dgpt} with SS+ & \textbf{R}: it depends on modular & \cmark & \cmark & \cmark & e \\ \hline
{\dgpt} with SS+ (o) & \textbf{R}: it depends on modular exponentiation & \cmark & \cmark & \cmark & e \\ \hline
Model  & \textbf{Q}:what was sadat seeking by releasing islamists from prison ? & \rotatebox[origin=c]{270}{correctness} & \rotatebox[origin=c]{270}{complete-sentence} & \rotatebox[origin=c]{270}{grammaticality} & \rotatebox[origin=c]{270}{majority option}  \\ \hline 
{\coqa} B. & \textbf{R}: political support in his struggle against leftists & \cmark & \cmark & \cmark & e \\ \hline
{\quac} B. & \textbf{R}: egyptian president anwar sadat – whose policies included opening egypt to western investment ( infitah ) ; transferring egypt 's allegiance from the soviet union to the united states & \xmark & \cmark & - & c \\ \hline
STs+BERT B. & \textbf{R}: sadat was seeking political support & \cmark & \cmark & \cmark & e \\ \hline
{\pgnp} with SS+ & \textbf{R}: it was making peace with israel & \xmark & \cmark & - & c \\ \hline
{\gpt} with SS+ & \textbf{R}: it was making peace with israel & \xmark & \cmark & - & c \\ \hline
{\gptp} with SS+ & \textbf{R}: sadat seeking was making peace with israel & \xmark & \cmark & - & c \\ \hline
{\dgpt} with SS+ & \textbf{R}: sadat was making peace with israel & \xmark & \cmark & - & c \\ \hline
{\dgpt} with SS+ (o) & \textbf{R}: sadat was seeking political support & \cmark & \cmark & \cmark & e \\ \hline
Model  & \textbf{Q}:how did the better jacksonville plan generate money ? & \rotatebox[origin=c]{270}{correctness} & \rotatebox[origin=c]{270}{complete-sentence} & \rotatebox[origin=c]{270}{grammaticality} & \rotatebox[origin=c]{270}{majority option}  \\ \hline 
{\coqa} B. & \textbf{R}: authorized a half-penny sales tax & \cmark & \xmark & - & b \\ \hline
{\quac} B. & \textbf{R}: when a consolidation referendum was held in 1967 , voters approved the plan . on october 1 , 1968 , the governments merged to create the consolidated city of jacksonville & \xmark & \cmark & - & c \\ \hline
STs+BERT B. & \textbf{R}: it generated money authorized a half-penny sales tax & \cmark & \cmark & \xmark & d \\ \hline
{\pgnp} with SS+ & \textbf{R}: it generated money half-penny sales tax & \cmark & \cmark & \xmark & d \\ \hline
{\gpt} with SS+ & \textbf{R}: it generated money half-penny sales tax & \cmark & \cmark & \xmark & d \\ \hline
{\gptp} with SS+ & \textbf{R}: it generated half-penny sales tax money & \cmark & \cmark & \cmark & e \\ \hline
{\dgpt} with SS+ & \textbf{R}: it generated money for half-penny sales tax & \cmark & \cmark & \cmark & e \\ \hline
{\dgpt} with SS+ (o) & \textbf{R}: it generated money authorized a half-penny sales tax & \cmark & \cmark & \xmark & d \\ \hline
Model  & \textbf{Q}:how many items is the university library home to ? & \rotatebox[origin=c]{270}{correctness} & \rotatebox[origin=c]{270}{complete-sentence} & \rotatebox[origin=c]{270}{grammaticality} & \rotatebox[origin=c]{270}{majority option}  \\ \hline 
{\coqa} B. & \textbf{R}: over two million & \cmark & \xmark & - & b \\ \hline
{\quac} B. & \textbf{R}: another important library – the university library , founded in 1816 , is home to over two million items & \cmark & \cmark & \cmark & e \\ \hline
STs+BERT B. & \textbf{R}: it to is over two million & \cmark & \cmark & \xmark & d \\ \hline
{\pgnp} with SS+ & \textbf{R}: it is home to over two million & \cmark & \cmark & \cmark & e \\ \hline
{\gpt} with SS+ & \textbf{R}: it is home to over two million & \cmark & \cmark & \cmark & e \\ \hline
{\gptp} with SS+ & \textbf{R}: it is home to over two million & \cmark & \cmark & \cmark & e \\ \hline
{\dgpt} with SS+ & \textbf{R}: it to is over two million & \cmark & \cmark & \xmark & d \\ \hline
{\dgpt} with SS+ (o) & \textbf{R}: it to is over two million & \cmark & \cmark & \xmark & d \\ \hline
Model  & \textbf{Q}:which sea was oil discovered in ? & \rotatebox[origin=c]{270}{correctness} & \rotatebox[origin=c]{270}{complete-sentence} & \rotatebox[origin=c]{270}{grammaticality} & \rotatebox[origin=c]{270}{majority option}  \\ \hline 
{\coqa} B. & \textbf{R}: North Sea & \cmark & \xmark & - & b \\ \hline
{\quac} B. & \textbf{R}: " it \ ' s scotland \ ' s oil " campaign of the scottish national party ( snp ) & \xmark & \xmark & - & a \\ \hline
LGRs+BERT B. & \textbf{R}: oil was discovered in north & \cmark & \cmark & \xmark & d \\ \hline
{\pgnp} with SS+ & \textbf{R}: oil was discovered in the north sea & \cmark & \cmark & \cmark & e \\ \hline
{\gpt} with SS+ & \textbf{R}: oil was discovered in the north sea & \cmark & \cmark & \cmark & e \\ \hline
{\gptp} with SS+ & \textbf{R}: it was discovered in the north sea & \cmark & \cmark & \cmark & e \\ \hline
{\dgpt} with SS+ & \textbf{R}: it was discovered in the north sea & \cmark & \cmark & \cmark & e \\ \hline
{\dgpt} with SS+ (o) & \textbf{R}: oil was discovered in north & \cmark & \cmark & \xmark & d \\ \hline
Model  & \textbf{Q}:where are jersey and guernsey & \rotatebox[origin=c]{270}{correctness} & \rotatebox[origin=c]{270}{complete-sentence} & \rotatebox[origin=c]{270}{grammaticality} & \rotatebox[origin=c]{270}{majority option}  \\ \hline 
{\coqa} B. & \textbf{R}: Channel Islands & \cmark & \xmark & - & b \\ \hline
{\quac} B. & \textbf{R}: the customary law of normandy was developed between the 10th and 13th centuries and survives today through the legal systems of jersey and guernsey in the channel islands & \cmark & \cmark & \cmark & e \\ \hline
LGRs+BERT B. & \textbf{R}: they are in channel islands & \cmark & \cmark & \cmark & e \\ \hline
{\pgnp} with SS+ & \textbf{R}: they are in the channel islands & \cmark & \cmark & \cmark & e \\ \hline
{\gpt} with SS+ & \textbf{R}: they are on the channel islands & \cmark & \cmark & \cmark & e \\ \hline
{\gptp} with SS+ & \textbf{R}: they are on the channel islands & \cmark & \cmark & \cmark & e \\ \hline
{\dgpt} with SS+ & \textbf{R}: they are in the channel islands & \cmark & \cmark & \cmark & e \\ \hline
{\dgpt} with SS+ (o) & \textbf{R}: they are in channel islands & \cmark & \cmark & \cmark & e \\ \hline
Model  & \textbf{Q}:near chur , which direction does the rhine turn ? & \rotatebox[origin=c]{270}{correctness} & \rotatebox[origin=c]{270}{complete-sentence} & \rotatebox[origin=c]{270}{grammaticality} & \rotatebox[origin=c]{270}{majority option}  \\ \hline 
{\coqa} B. & \textbf{R}: north & \cmark & \xmark & - & b \\ \hline
{\quac} B. & \textbf{R}: near tamins-reichenau the anterior rhine and the posterior rhine join and form the rhine & \xmark & \cmark & - & c \\ \hline
LGRs+BERT B. & \textbf{R}: it turns north & \cmark & \cmark & \cmark & e \\ \hline
{\pgnp} with SS+ & \textbf{R}: it turns north & \cmark & \cmark & \cmark & e \\ \hline
{\gpt} with SS+ & \textbf{R}: it turns north & \cmark & \cmark & \cmark & e \\ \hline
{\gptp} with SS+ & \textbf{R}: it turns to the north & \cmark & \cmark & \cmark & e \\ \hline
{\dgpt} with SS+ & \textbf{R}: it turns north & \cmark & \cmark & \cmark & e \\ \hline
{\dgpt} with SS+ (o) & \textbf{R}: it turns north & \cmark & \cmark & \cmark & e \\ \hline
Model  & \textbf{Q}:what kind of contract is given when the contractor is given a performance specification and must undertake the project from design to construction , while adhering to the performance specifications ? & \rotatebox[origin=c]{270}{correctness} & \rotatebox[origin=c]{270}{complete-sentence} & \rotatebox[origin=c]{270}{grammaticality} & \rotatebox[origin=c]{270}{majority option}  \\ \hline 
{\coqa} B. & \textbf{R}: design build" contract & \cmark & \xmark & - & b \\ \hline
{\quac} B. & \textbf{R}: the modern trend in design is toward integration of previously separated specialties , especially among large firms & \xmark & \cmark & - & c \\ \hline
LGRs+BERT B. & \textbf{R}: a "design build" contract is given & \cmark & \cmark & \cmark & e \\ \hline
{\pgnp} with SS+ & \textbf{R}: design build is given & \cmark & \cmark & \cmark & e \\ \hline
{\gpt} with SS+ & \textbf{R}: the design build is given & \cmark & \cmark & \cmark & e \\ \hline
{\gptp} with SS+ & \textbf{R}: design build is given a performance specification and must undertake the project & \cmark & \cmark & \cmark & e \\ \hline
{\dgpt} with SS+ & \textbf{R}: design build is given & \cmark & \cmark & \cmark & e \\ \hline
{\dgpt} with SS+ (o) & \textbf{R}: the " design build " contract is given & \cmark & \cmark & \cmark & e \\ \hline
Model  & \textbf{Q}:how many protestants live in france today ? & \rotatebox[origin=c]{270}{correctness} & \rotatebox[origin=c]{270}{complete-sentence} & \rotatebox[origin=c]{270}{grammaticality} & \rotatebox[origin=c]{270}{majority option}  \\ \hline 
{\coqa} B. & \textbf{R}: Approximately one million & \cmark & \xmark & - & b \\ \hline
{\quac} B. & \textbf{R}: approximately one million protestants in modern france represent some 2 \% of its population & \cmark & \cmark & \cmark & e \\ \hline
LGRs+BERT B. & \textbf{R}: one million live in france today & \cmark & \cmark & \cmark & e \\ \hline
{\pgnp} with SS+ & \textbf{R}: one million live in france today & \cmark & \cmark & \cmark & e \\ \hline
{\gpt} with SS+ & \textbf{R}: one million live in france today & \cmark & \cmark & \cmark & e \\ \hline
{\gptp} with SS+ & \textbf{R}: one million live in france today & \cmark & \cmark & \cmark & e \\ \hline
{\dgpt} with SS+ & \textbf{R}: one million live in france today & \cmark & \cmark & \cmark & e \\ \hline
{\dgpt} with SS+ (o) & \textbf{R}: one million live in france today & \cmark & \cmark & \cmark & e \\ \hline
Model  & \textbf{Q}:what is raghuram rajan 's career ? & \rotatebox[origin=c]{270}{correctness} & \rotatebox[origin=c]{270}{complete-sentence} & \rotatebox[origin=c]{270}{grammaticality} & \rotatebox[origin=c]{270}{majority option}  \\ \hline 
{\coqa} B. & \textbf{R}: Central Banking economist & \cmark & \xmark & - & b \\ \hline
{\quac} B. & \textbf{R}: central banking economist raghuram rajan argues that " systematic economic inequalities & \cmark & \xmark & - & b \\ \hline
LGRs+BERT B. & \textbf{R}: he is economist & \cmark & \cmark & \xmark & d \\ \hline
{\pgnp} with SS+ & \textbf{R}: it is central banking economist & \cmark & \cmark & \cmark & e \\ \hline
{\gpt} with SS+ & \textbf{R}: it is central banking economist & \cmark & \cmark & \cmark & e \\ \hline
{\gptp} with SS+ & \textbf{R}: it is central banking economist & \cmark & \cmark & \cmark & e \\ \hline
{\dgpt} with SS+ & \textbf{R}: it is central banking economist & \cmark & \cmark & \cmark & e \\ \hline
{\dgpt} with SS+ (o) & \textbf{R}: he is economist & \cmark & \cmark & \xmark & d \\ \hline
Model  & \textbf{Q}:what type of steam engines produced most power up to the early 20th century ? & \rotatebox[origin=c]{270}{correctness} & \rotatebox[origin=c]{270}{complete-sentence} & \rotatebox[origin=c]{270}{grammaticality} & \rotatebox[origin=c]{270}{majority option}  \\ \hline 
{\coqa} B. & \textbf{R}: Reciprocating piston type steam engines & \cmark & \xmark & - & b \\ \hline
{\quac} B. & \textbf{R}: reciprocating piston type steam engines remained the dominant source of power until the early 20th century , when advances in the design of electric motors and internal combustion engines & \cmark & \cmark & \cmark & e \\ \hline
LGRs+BERT B. & \textbf{R}: reciprocating piston produced most power up & \cmark & \cmark & \xmark & d \\ \hline
{\pgnp} with SS+ & \textbf{R}: reciprocating piston type produced most power up & \cmark & \cmark & \xmark & d \\ \hline
{\gpt} with SS+ & \textbf{R}: reciprocating piston type produced most power up & \cmark & \cmark & \xmark & d \\ \hline
{\gptp} with SS+ & \textbf{R}: the reciprocating piston type produced most power up to the early 20th century & \cmark & \cmark & \cmark & e \\ \hline
{\dgpt} with SS+ & \textbf{R}: reciprocating piston type produced most power up to the early 20th century & \cmark & \cmark & \cmark & e \\ \hline
{\dgpt} with SS+ (o) & \textbf{R}: reciprocating piston produced most power up to the early 20th century & \cmark & \cmark & \cmark & e \\ \hline
Model  & \textbf{Q}:where did france win a war in the 1950 's & \rotatebox[origin=c]{270}{correctness} & \rotatebox[origin=c]{270}{complete-sentence} & \rotatebox[origin=c]{270}{grammaticality} & \rotatebox[origin=c]{270}{majority option}  \\ \hline 
{\coqa} B. & \textbf{R}: Algeria & \cmark & \xmark & - & b \\ \hline
{\quac} B. & \textbf{R}: france fought and lost a bitter war in vietnam in the 1950s & \xmark & \cmark & - & c \\ \hline
LGRs+BERT B. & \textbf{R}: france won a war in the 1950 's algeria & \cmark & \cmark & \cmark & e \\ \hline
{\pgnp} with SS+ & \textbf{R}: france won a war in vietnam & \xmark & \cmark & - & c \\ \hline
{\gpt} with SS+ & \textbf{R}: france won a war in vietnam & \xmark & \cmark & - & c \\ \hline
{\gptp} with SS+ & \textbf{R}: france won a war in vietnam & \xmark & \cmark & - & c \\ \hline
{\dgpt} with SS+ & \textbf{R}: france won a war in vietnam & \xmark & \cmark & - & c \\ \hline
{\dgpt} with SS+ (o) & \textbf{R}: france won a war in algeria & \cmark & \cmark & \cmark & e \\ \hline
Model  & \textbf{Q}:who did the ottoman empire ally with in ww i ? & \rotatebox[origin=c]{270}{correctness} & \rotatebox[origin=c]{270}{complete-sentence} & \rotatebox[origin=c]{270}{grammaticality} & \rotatebox[origin=c]{270}{majority option}  \\ \hline 
{\coqa} B. & \textbf{R}: Germany & \cmark & \xmark & - & b \\ \hline
{\quac} B. & \textbf{R}: the ottoman empire gradually declined into the late nineteenth century . the empire allied with germany & \cmark & \cmark & \cmark & e \\ \hline
LGRs+BERT B. & \textbf{R}: germany did the ottoman empire ally with in ww i & \cmark & \cmark & \xmark & d \\ \hline
{\pgnp} with SS+ & \textbf{R}: it separated with germany & \xmark & \cmark & - & c \\ \hline
{\gpt} with SS+ & \textbf{R}: it allyed with germany & \cmark & \cmark & \cmark & e \\ \hline
{\gptp} with SS+ & \textbf{R}: it allyed with germany & \cmark & \cmark & \cmark & e \\ \hline
{\dgpt} with SS+ & \textbf{R}: it allied germany & \cmark & \cmark & \xmark & d \\ \hline
{\dgpt} with SS+ (o) & \textbf{R}: it allied germany & \cmark & \cmark & \xmark & d \\ \hline
Model  & \textbf{Q}:when was ambulatory care pharmacy approved as its own certification ? & \rotatebox[origin=c]{270}{correctness} & \rotatebox[origin=c]{270}{complete-sentence} & \rotatebox[origin=c]{270}{grammaticality} & \rotatebox[origin=c]{270}{majority option}  \\ \hline 
{\coqa} B. & \textbf{R}: In 2011 & \cmark & \xmark & - & b \\ \hline
{\quac} B. & \textbf{R}: in 2011 the board of pharmaceutical specialties approved ambulatory care pharmacy practice as a separate board certification & \cmark & \cmark & \cmark & e \\ \hline
LGRs+BERT B. & \textbf{R}: it was approved in 2011 & \cmark & \cmark & \cmark & e \\ \hline
{\pgnp} with SS+ & \textbf{R}: it was approved in 2011 & \cmark & \cmark & \cmark & e \\ \hline
{\gpt} with SS+ & \textbf{R}: it was approved in 2011 & \cmark & \cmark & \cmark & e \\ \hline
{\gptp} with SS+ & \textbf{R}: it was approved in 2011 & \cmark & \cmark & \cmark & e \\ \hline
{\dgpt} with SS+ & \textbf{R}: it was approved in 2011 & \cmark & \cmark & \cmark & e \\ \hline
{\dgpt} with SS+ (o) & \textbf{R}: it was approved in 2011 & \cmark & \cmark & \cmark & e \\ \hline
Model  & \textbf{Q}:when did arpnet and sita become operational & \rotatebox[origin=c]{270}{correctness} & \rotatebox[origin=c]{270}{complete-sentence} & \rotatebox[origin=c]{270}{grammaticality} & \rotatebox[origin=c]{270}{majority option}  \\ \hline 
{\coqa} B. & \textbf{R}: 1969 & \cmark & \xmark & - & b \\ \hline
{\quac} B. & \textbf{R}: arpanet and sita hln became operational in 1969 & \cmark & \cmark & \cmark & e \\ \hline
LGRs+BERT B. & \textbf{R}: 1969 & \cmark & \xmark & - & b \\ \hline
{\pgnp} with SS+ & \textbf{R}: they became operational in 1969 & \cmark & \cmark & \cmark & e \\ \hline
{\gpt} with SS+ & \textbf{R}: they became operational in 1969 & \cmark & \cmark & \cmark & e \\ \hline
{\gptp} with SS+ & \textbf{R}: they became operational in 1969 & \cmark & \cmark & \cmark & e \\ \hline
{\dgpt} with SS+ & \textbf{R}: they became operational in 1969 & \cmark & \cmark & \cmark & e \\ \hline
{\dgpt} with SS+ (o) & \textbf{R}: they became operational in 1969 & \cmark & \cmark & \cmark & e \\ \hline
Model  & \textbf{Q}:how much did saudi arabia spend on spreading wahhabism ? & \rotatebox[origin=c]{270}{correctness} & \rotatebox[origin=c]{270}{complete-sentence} & \rotatebox[origin=c]{270}{grammaticality} & \rotatebox[origin=c]{270}{majority option}  \\ \hline 
{\coqa} B. & \textbf{R}: over 100 billion dollars & \cmark & \xmark & - & b \\ \hline
{\quac} B. & \textbf{R}: saudi arabia spent over 100 billion dollars in the ensuing decades for helping spread its fundamentalist interpretation of islam & \cmark & \cmark & \cmark & e \\ \hline
LGRs+BERT B. & \textbf{R}: saudi arabia spent over 100 billion dollars & \cmark & \cmark & \cmark & e \\ \hline
{\pgnp} with SS+ & \textbf{R}: saudi arabia spent over 100 billion dollars & \cmark & \cmark & \cmark & e \\ \hline
{\gpt} with SS+ & \textbf{R}: saudi arabia spent over 100 billion dollars & \cmark & \cmark & \cmark & e \\ \hline
{\gptp} with SS+ & \textbf{R}: saudi arabia spent over 100 billion dollars & \cmark & \cmark & \cmark & e \\ \hline
{\dgpt} with SS+ & \textbf{R}: saudi arabia spent over 100 billion dollars & \cmark & \cmark & \cmark & e \\ \hline
{\dgpt} with SS+ (o) & \textbf{R}: saudi arabia spent over 100 billion dollars & \cmark & \cmark & \cmark & e \\ \hline
\end{longtable}

\end{document}